\documentclass[letterpaper, 10 pt, conference]{ieeeconf}  

\IEEEoverridecommandlockouts                              

\usepackage{graphicx}
\usepackage{cite}
\usepackage{amsmath,amssymb,amsfonts}
\usepackage{algorithm}
\usepackage{algorithmic}
\usepackage{comment}
\usepackage{caption}
\usepackage{subcaption}
\usepackage{multirow}
\usepackage{float}    
\usepackage{placeins} 
\usepackage{hyperref}
\usepackage{booktabs}
\usepackage{dsfont}
\usepackage{tikz}
\usepackage{xcolor}
\usepackage{amsmath}
\usetikzlibrary{arrows.meta,positioning,fit,calc,shapes.geometric}
\usepackage{bm}
\usepackage[acronym]{glossaries}

\newacronym{ad}{AD}{autonomous driving}
\newacronym{e2e}{E2E}{End-to-end}
\newacronym{il}{IL}{Imitation learning}
\newacronym{rl}{RL}{reinforcement learning}
\newacronym{pomdp}{POMDP}{Partially Observable Markov Decision Process}
\newacronym{iqn}{IQN}{Implicit quantile network}
\newacronym{tm}{TM}{Traffic Manager}
\newacronym{cvar}{CVaR}{Conditional Value at Risk}
\newacronym{cmdp}{CMDP}{Constrained Markov Decision Process}
\newacronym{cbfs}{CBFs}{Control Barrier Functions}

\usepackage{eso-pic}
\newcommand{\mycopyrighttext}{%
  \footnotesize
  \noindent
  \textcopyright~2026 IEEE. Personal use of this material is permitted. Permission from IEEE must be obtained for all other uses, in any current or future media, including reprinting/republishing this material for advertising or promotional purposes, creating new collective works, for resale or redistribution to servers or lists, or reuse of any copyrighted component of this work in other works.\\
  The IEEE International Conference on Intelligent Transportation Systems (ITSC 2026) - 15-18 September, 2026.
}

\AtBeginDocument{%
  \AddToShipoutPicture*{%
    \AtPageUpperLeft{%
      \put(55, -40){
        \parbox{\dimexpr\textwidth-4pt\relax}{\raggedright \mycopyrighttext}
      }%
    }
  }
}
\title{\LARGE \bf Uncertainty-Aware and Temporally Regulated Expert Advice \\ in Reinforcement Learning for Autonomous Driving}
\author{Ahmed Abouelazm$^{1,2}$, Felix Klingebiel$^{2}$, Philip Schoerner$^{1,2}$, and J. Marius Zöllner$^{1,2}$
\thanks{$^{1}$Authors are with the FZI Research Center for Information Technology, Germany {\tt\small name@fzi.de}}%
\thanks{$^{2}$Authors are with the Karlsruhe Institute of Technology, Germany}%
}

\begin{document}
\bstctlcite{IEEEexample:BSTcontrol}
\maketitle
\thispagestyle{empty}
\pagestyle{empty}

\begin{abstract}
    Exploration in reinforcement learning for autonomous driving is inherently unsafe: agents must experience novel behaviors to learn, yet exploration can lead to collisions or off-road driving. We propose an uncertainty-aware framework that leverages expert advice to guide exploration while avoiding long-term dependence. Advice is triggered when epistemic or aleatoric uncertainty exceeds adaptive thresholds derived from rolling buffers, ensuring advice evolves with the agent’s confidence. A commitment–cooldown strategy with a stochastic early-stop heuristic regulates the duration and frequency of guidance, exposing the agent to coherent maneuvers without exhausting the advice budget. Expert and agent experiences are combined in a shared replay buffer within an off-policy implicit quantile network (IQN) backbone, enabling efficient reuse of expert trajectories. Experiments in CARLA show that our method outperforms the IQN baseline, improving success by 5–7\% and reducing failures, demonstrating that risk-sensitive uncertainty coupled with regulated expert integration enables safer and more efficient exploration for sensor-based RL policy learning in unsignalized intersection navigation.
    
\end{abstract}
\section{Introduction}
\label{sec:Introduction}
\gls{e2e} has emerged as a promising paradigm for \gls{ad}, mapping raw sensory inputs directly to control actions and reducing reliance on handcrafted modular pipelines~\cite{chen2024end}. Among \gls{e2e} approaches, \gls{rl} is particularly appealing: unlike \gls{il}, which depends on expert demonstrations, \gls{rl} enables agents to adapt through direct interaction with the environment~\cite{sutton1998reinforcement}. By exploring different behaviors and receiving feedback via rewards, \gls{rl} agents can handle complex conditions such as occlusions, erratic drivers, and rare events that are underrepresented in datasets.

However, \gls{rl} also poses fundamental challenges in sample efficiency and safety~\cite{gu2024review}. The learning process is driven by the interplay between exploration, where the agent deliberately samples novel actions to acquire new knowledge about the environment, and exploitation, where it applies its current policy to maximize expected rewards~\cite{sutton1998reinforcement}. Exploration is indispensable for avoiding suboptimal policies and uncovering effective driving strategies, yet it inherently involves unsafe or undesirable behaviors~\cite{wachi2023safe}.

At the same time, extensive exploration leads to poor sample efficiency, as agents require a vast number of interactions before converging to a reliable policy~\cite{dulac2021challenges}. These limitations underscore the need for mechanisms that constrain or prune exploration, directing the agent toward informative regions of the state space while preventing unnecessary behaviors.

\textbf{Research Gap. } Existing work on safe exploration in \gls{rl} for \gls{ad} has followed several directions. Formal approaches such as \gls{cmdp}~\cite{hu2024long}, and Lyapunov functions~\cite{liu2024automatic} 
encode safety explicitly, but they require handcrafted risk definitions, involve complex optimization, and provide no mechanism to guide policies toward safer actions. To address these limitations, expert knowledge has been introduced through demonstrations. Demonstration-based pre-training~\cite{choi2025enhancing} accelerates convergence yet leaves agents vulnerable to unsafe exploration and provides no corrective feedback, while human-in-the-loop systems~\cite{huang2024human} allow targeted interventions but are not scalable and remain detached from the agent’s internal decision process. 

More scalable alternatives replace humans with rule-based~\cite{wang2023safe} or learned~\cite{xue2023guarded} expert policies, which intervene automatically based on predefined heuristics. Although effective at preventing unsafe rollouts, rule-based triggers are overly conservative and context-dependent, while learned experts assume strong generalization and may overwrite valid actions with poor ones when this assumption fails. Recent advances shift control to the agent, allowing it to query expert advice when needed via triggers such as state uncertainty~\cite{da2020uncertainty} and state novelty~\cite{ilhan2021student}. However, these approaches remain largely state-centric, neglect the risk induced by the agent’s exploratory actions, and lack mechanisms to regulate the frequency and impact of advice.

To address these shortcomings, methods are needed that extend uncertainty estimation to capture action-related risk, combined with adaptive mechanisms that regulate advice and embed safety into the learned policy.

\textbf{Contribution.} This work proposes an \gls{rl} framework for \gls{ad} that improves exploration safety by extending uncertainty estimation to account for action-related risk and integrating expert input through adaptive mechanisms. The key contributions are:

\begin{itemize}
\item \textbf{Uncertainty-aware expert guidance:} the agent requests advice in states with high epistemic or aleatoric uncertainty, storing both expert and agent transitions in a shared buffer for training.
\item \textbf{Uncertainty estimation:} aleatoric uncertainty (environmental risk) is derived from return variance, while epistemic uncertainty (limited knowledge) is measured via bootstrapped ensembles, enabling detection of risky and underexplored states.
\item \textbf{Commitment–cooldown strategy:} regulates advice use by applying expert actions over short horizons and enforcing cooldowns, ensuring effective guidance without overreliance.
\end{itemize}

\section{related work}
Safe \gls{rl} in \gls{ad} is challenging due to the need to balance efficiency with safety requirements. A common approach is to regulate exploration through constraints or learned objectives. \gls{cmdp} formulations constrain cumulative risk below predefined thresholds~\cite{wang2023autonomous,hu2024long}, while Lyapunov-based methods enforce stability by requiring a Lyapunov function to decrease along trajectories~\cite{liu2024automatic,du2023reinforcement}. \gls{cbfs} define forward-invariant safe sets and enforce constraints to keep trajectories within these sets~\cite{tan2023value,tan2024safe}, and distributional RL optimizes risk-sensitive returns such as \gls{cvar}~\cite{bernhard2019addressing,kamran2021minimizing}.

Despite their differences, these approaches share key limitations: they solve complex optimization problems, depend on manually defined costs or stability conditions, or rely on learned objectives that are especially noisy at the start of training, and crucially, they lack mechanisms to actively guide the policy toward safer actions. This prompted the use of expert knowledge and demonstrations, which provide more direct and effective signals during exploration.

Several works utilize demonstrations to pre-train policies, which accelerates convergence but leaves the agent vulnerable to unsafe exploration and provides no corrective feedback during training~\cite{choi2025enhancing,pfeiffer2018reinforced}. Early integration strategies alternate between collecting episodes from the expert and from the agent’s policy~\cite{gao2025ril,chekroun2023gri}, which improves robustness but still limits expert input to entire episodes rather than targeted interventions. As these episodes predominantly capture nominal driving and seldom illustrate recovery from unsafe states, these methods fail to teach agents how to act in the situations where guidance is most needed. This limitation has driven interest in more active systems, where experts can intervene, or the agent selectively queries advice.

A common form of active expert intervention is human-in-the-loop training~\cite{peng2023learning,huang2024human}, where a supervisor monitors the environment and provides corrective actions to guide the agent out of unsafe states. While effective, this approach is not scalable for long training, introduces bias in when and how interventions occur, and remains detached from the agent’s decision process, as humans cannot fully observe the agent’s intentions and may override potential recovery actions prematurely. These challenges have shifted attention toward alternative expert policies that can offer more scalable and consistent intervention during training.

For such alternative policies, explicit intervention signals are needed to decide when to prune unsafe exploration. One direction relies on rule-based definitions of risk, where interventions occur if the risk associated with the agent’s behavior exceeds a threshold~\cite{mo2021safe,kamran2021minimizing} or if the behavior is identified as unsafe through safety analysis~\cite{wang2023safe,thumm2023reducing}. While effective at avoiding unsafe rollouts, these definitions are often complex, context-dependent, and can restrict exploration and hinder policy improvement.

To overcome the rigidity of rule-based experts, recent works employ learned expert policies. These experts intervene if the agent's action has a low likelihood under the expert policy~\cite{peng2022safe}, or the expert’s value function deems it unsafe~\cite{wang2023learning}, or it deviates significantly from the expert’s optimal action~\cite{xue2023guarded}. Although more flexible, these methods assume that the expert policy generalizes reliably across the state space. When this assumption fails, interventions may overwrite reasonable agent actions with poor ones, leading to suboptimal behavior. This has opened the way for approaches where the agent selectively queries expert advice instead of being forced into intervention.

Expert advice gives the agent more control by allowing it to decide when to query the expert. Several works trigger advice based on signals derived from the agent’s learning process, such as state uncertainty~\cite{kelly2019hg,da2020uncertainty}, state novelty~\cite{ilhan2021student,wei2025agent}, or similarity to unsafe states~\cite{bethell2024safe}. While these methods provide a more active framework, they focus primarily on state-based measures, neglect the risk induced by the agent’s actions, lack mechanisms to regulate the impact or frequency of advice, and often rely on fixed thresholds to decide when to intervene. These gaps motivate our framework for selective expert advice, which expands uncertainty estimation to explicitly account for action-related risks and employs adaptive triggering to deliver targeted guidance that is internalized by the policy during training.
\section{Methodology}
\begin{figure*}[!t]
\centering
\includegraphics[width=\linewidth]{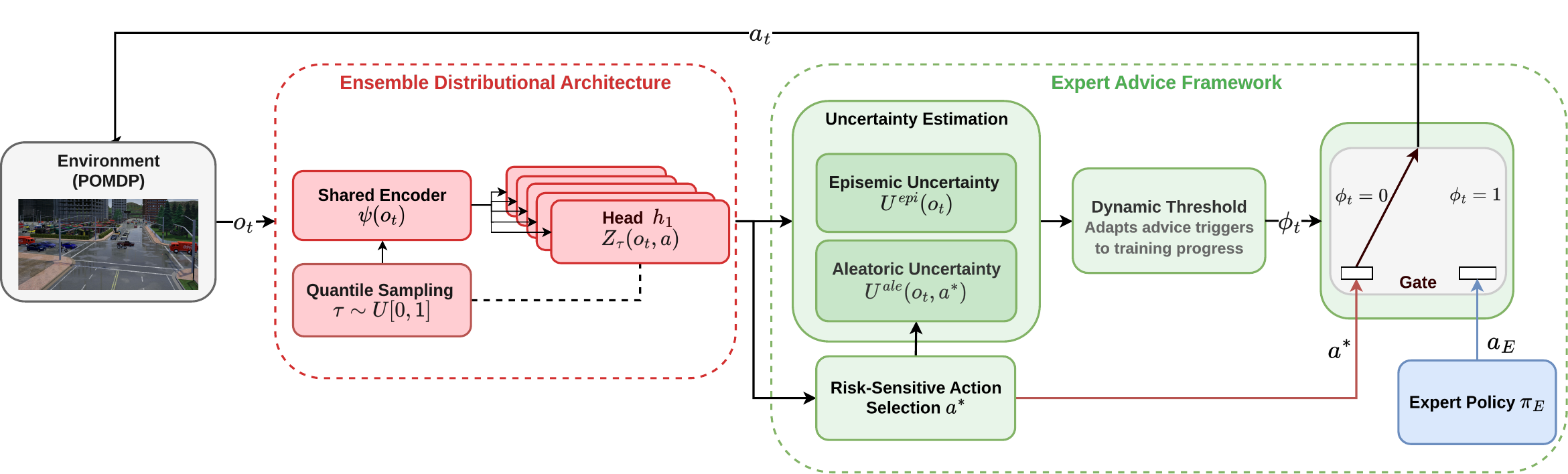}
\caption{Overview of the proposed uncertainty-aware expert guidance framework. 
An ensemble distributional architecture provides epistemic and aleatoric 
uncertainty estimates. The agent queries the expert whenever either uncertainty 
exceeds adaptive thresholds derived from rolling buffers. A commitment–cooldown 
mechanism regulates the frequency and duration of advice.}
\label{fig:advice_framework}
\vspace{-0.4cm}
\end{figure*}

Our framework, illustrated in Fig.~\ref{fig:advice_framework}, introduces an advice mechanism in which the agent requests guidance when its policy is uncertain about the state (epistemic uncertainty) or potentially unsafe due to its actions (aleatoric uncertainty). An adaptive triggering strategy regulates when and for how long advice is applied, ensuring that interventions accelerate learning without fostering long-term dependence. 
\subsection{Problem Formulation}
\label{sec:ProblemFormulation}
We formalize policy learning for \gls{ad} as a \gls{pomdp}, defined by the tuple
$\mathcal{M} = \langle \mathcal{S}, \mathcal{A}, \mathcal{P}, r, \mathcal{O}, \gamma \rangle$
where $\mathcal{S}$ and $\mathcal{A}$ denote the state and action spaces, and $\mathcal{P}(s' \mid s,a)$ the transition model describing the probability of reaching state $s'$ from state $s$ under action $a$. The reward $r(s,a)$ specifies the immediate feedback for taking action $a$ in state $s$, while $\gamma \in [0,1)$ is the discount factor that balances short- and long-term returns. In contrast to an MDP, the agent cannot directly access the underlying state $s_t$. Instead, after executing an action $a_t$, it receives an observation $o_{t+1} \sim \mathcal{O}(s_{t+1}, a_t)$ which is generated from the hidden state $s_{t+1}$ via a sensor model. 

Beyond its own policy $\pi_\theta$, the agent has access to an expert policy $\pi_E$ that can be queried for advice, subject to a global budget $B$ that limits the total number of requests and reflects practical constraints such as limited availability of human supervision or computationally expensive expert controllers. We treat the expert as a black-box oracle and make no assumptions about its optimality or consistency; we only require that it can provide an action for a given observation or state. To regulate when advice is requested, we introduce an advice decision policy $\phi: \mathcal{O} \times \mathcal{A} \to \{0,1\}$. For convenience, we denote $\phi_t = \phi(o_t, a_t)$, where $\phi_t = 1$ indicates that the expert policy $\pi_E$ is queried at time $t$, and $\phi_t = 0$ denotes that the agent 
follows its own policy $\pi_\theta$.
\subsection{Policy Learning Algorithm}
Our framework requires a learning algorithm that can integrate expert advice with agent experiences. This motivates an off-policy approach, where experiences are collected under a behavior policy $\pi_b$ and used to optimize the target policy $\pi_\theta$. This decoupling allows expert actions to be inserted into a shared replay buffer alongside agent rollouts, ensuring that advice can be reused across multiple updates. In contrast, on-policy methods couple data collection and optimization, so expert overrides produce biased gradient estimates. Additionally, these methods discard rollouts after each update; expert corrections cannot be reused, and their effect remains short-lived.

Partial observability further complicates \gls{ad}, as decisions must be made under both epistemic (limited knowledge) and aleatoric (sensor noise and occlusions) uncertainty. Conventional value-based methods optimize only expected returns, which do not capture these uncertainties. To address this, we adopt a distributional \gls{rl} formulation that models the entire return distribution.  

Specifically, we employ the \gls{iqn}~\cite{dabney2018implicit}, which uniformly samples quantile fractions $\tau \sim \mathcal{U}[0,1]$. Each $\tau$ is first embedded as cosine features and combined with the observation latent representation. The network then uses this joint representation to output $Z_\tau(o_t,a)$, which approximates the $\tau$-quantile of the return distribution, i.e., the inverse cumulative distribution function $F^{-1}_{Z(o_t,a)}(\tau)$, as shown in Eq.~\ref{eq:inverse_cdf}. Furthermore, \gls{iqn} supports risk-sensitive decision-making by conditioning on specific regions of the distribution, for instance, focusing on lower quantiles to prioritize safer driving.
\begin{equation}
    Z_{\tau}(o_t,a) \; \approx \; F^{-1}_{\,Z(o_t,a)}(\tau), \quad \tau \sim \mathcal{U}[0,1]
    \label{eq:inverse_cdf}
\end{equation}

\subsection{Uncertainty Estimation}
\label{sec:uncertainty}
Uncertainty is central to our proposed approach, as it dictates when the agent should request expert advice, enabling dynamic adaptation in high-risk or ambiguous scenarios. To efficiently estimate uncertainty, we adopt an ensemble framework with multiple heads sharing a latent encoder, as illustrated in Fig.~\ref{fig:advice_framework}. Each head is trained on a different subset of the replay buffer~\cite {osband2016deep}, ensuring that variability in underexplored or ambiguous regions of the state space is reflected in the ensemble’s estimates. 

The model output is structured as $N_H \times N_\tau \times N_\mathcal{A}$, where $N_H$ denotes the number of ensemble heads, $N_\tau$ the sampled quantiles, and $N_\mathcal{A}$ the available actions. This output captures variability across ensemble predictions, distributional samples, and actions, and serves as the foundation for the advice mechanism that determines when expert intervention is desired. To ensure reliable decision-making, the agent must account for both epistemic and aleatoric uncertainty, the two principal sources of uncertainty in \gls{rl}~\cite{lockwood2022review}.
\subsubsection{Epistemic Uncertainty}
arises from the agent’s limited knowledge about the environment and is most pronounced in regions of the state space that have not been sufficiently explored~\cite{lakshminarayanan2017simple}. During training, this uncertainty gradually decreases as the agent collects more diverse experiences, enabling it to form more confident estimates in previously unfamiliar situations. For \gls{ad}, this is particularly relevant when the vehicle encounters uncommon road layouts or atypical maneuvers from other drivers. In such cases, high epistemic uncertainty reflects limited prior exposure but also indicates that the agent has the potential to improve its understanding if similar scenarios are encountered again.

In this work, we propose a state-based formulation that jointly considers the return distributions of all actions. This choice provides a more stable estimate and aligns with the definition of epistemic uncertainty as a lack of knowledge about the full state dynamics, rather than about individual actions in isolation. Concretely, we measure the epistemic uncertainty as the variability of the quantile function across ensemble heads, aggregated over all actions. 

To better leverage the expressive power of distributional \gls{rl}, we investigate two plausible formulations of epistemic uncertainty, each highlighting different aspects of the return distribution. Our objective is to assess which provides a more reliable signal in practice. The first formulation adopts a distributional perspective by comparing entire return distributions across ensemble heads using the Wasserstein distance $W_1$. For two heads $h$ and $h'$, the distance is computed by averaging absolute differences across quantile samples, as shown in Eq.~\ref{eq:wass_dist}, where $Z^{(h)}_{\tau_i}(o_t,a)$ denotes the $\tau_i$-quantile predicted by head $h$. 
\begin{equation}
W_1^{(h,h')}(o_t,a) = \dfrac{1}{N_\tau}\sum_{\tau_i \in \tau}
\left|Z^{(h)}_{\tau_i}(o_t,a)-Z^{(h')}_{\tau_i}(o_t,a)\right|
\label{eq:wass_dist}
\end{equation}
Epistemic uncertainty $U^{\text{epi}}_{\text{W}}$ is computed as the mean pairwise Wasserstein distance across ensemble head pairs, capturing variability, and aggregated over actions into a state-based measure, as illustrated in Eq.~\ref{eq:epi_wass}.
\begin{equation}
U^{\text{epi}}_{\text{W}}(o_t)
=\dfrac{1}{N_\mathcal{A}}\sum_{a\in \mathcal{A}} \dfrac{1}{\binom{N_H}{2}}
\sum_{\substack{h,h'=1 \\ h<h'}}^{N_H}
W_1^{(h,h')}(o_t,a)
\label{eq:epi_wass}
\end{equation}
The second formulation emphasizes a risk-sensitive perspective. Rather than comparing full distributions, we approximate the expected return using the \gls{cvar}~\cite{dabney2018implicit}, which focuses on the lower quantiles $\widetilde{\tau}$ determined by the risk parameter $\alpha$. For head $h$, the \gls{cvar} estimate is defined in Eq.~\ref{eq:cvar_dist}, where $Z^{(h)}_{\widetilde{\tau}_i}(o_t,a)$ denotes the predicted return at quantile $\widetilde{\tau}_i$.
\begin{equation}
\operatorname{CVaR}^{(h)}_\alpha(o_t,a) 
= \dfrac{1}{N_\tau}\sum_{\tau_i \in \tau} Z^{(h)}_{\widetilde{\tau}_i}(o_t,a),
\quad \widetilde{\tau}_i = \alpha \, \tau_i
\label{eq:cvar_dist}
\end{equation}
Epistemic uncertainty $U^{\text{epi}}_{\text{CVaR}}$ is then computed as the variance of these \gls{cvar} estimates across ensemble heads, aggregated over actions, as given in 
Eq.~\ref{eq:epi_cvar}.
\begin{equation}
U^{\text{epi}}_{\text{CVaR}}(o_t)
=\frac{1}{N_\mathcal{A}}\sum_{a\in \mathcal{A}}
\mathrm{Var}_H\!\big(\operatorname{CVaR}^{(h)}_\alpha(o_t,a)\big)
\label{eq:epi_cvar}
\end{equation}

\subsubsection{Aleatoric Uncertainty} reflects the inherent randomness in the environment and cannot be eliminated through training. It arises from factors such as sensor noise or partial observability~\cite{lockwood2022review}. For instance, when a vehicle approaches an occluded intersection, the presence, speed, or intention of unseen vehicles is fundamentally uncertain from the agent’s perspective, regardless of the agent’s prior experience. 

In this work, we follow the distributional interpretation used in~\cite{hoel2023ensemble}, where aleatoric uncertainty is derived from the spread of the predicted return distribution. To align this with our advice mechanism, uncertainty is evaluated only for the action the agent intends to execute, since this action determines whether intervention is required. Accordingly, aleatoric uncertainty is defined as the lower-tail return-distribution variance of the selected risk-sensitive action $a^*$, chosen by the greedy \gls{cvar}-based policy in Eq.~\ref{eq:greedy_policy}.
\begin{equation}
a^* = \operatorname*{arg\,max}_{a \in \mathcal{A}} \dfrac{1}{N_H} 
\sum_{h=1}^{N_H} \operatorname{CVaR}^{(h)}_\alpha(o_t,a)
\label{eq:greedy_policy}
\end{equation}
To quantify uncertainty for $a^*$, we focus on the lower-tail quantiles $\widetilde{\tau}$, which are determined by the risk parameter $\alpha$. For each quantile $\widetilde{\tau}_i$, we average the corresponding predictions across ensemble heads, yielding the ensemble-mean quantile estimate $\mu_{\widetilde{\tau}_i}$, as shown in Eq.~\ref{eq:ale_mean}. Aleatoric uncertainty is then computed as the variance of these ensemble-mean estimates over the lower-tail quantiles $\widetilde{\tau}$ (Eq.~\ref{eq:aleo_var}).
\begin{align}
    \mu_{\widetilde{\tau}_i}(o_t,a^*) &=
    \dfrac{1}{N_H}\sum_{h=1}^{N_H}
    Z^{(h)}_{\widetilde{\tau}_i}(o_t,a^*)
    \label{eq:ale_mean} \\
    U^{\text{ale}}_{\alpha}(o_t,a^*) &=
    \operatorname{Var}_{\widetilde{\tau}_i \in \widetilde{\tau}}
    \left[
    \mu_{\widetilde{\tau}_i}(o_t,a^*)
    \right]
    \label{eq:aleo_var}
\end{align}

\subsection{Expert Advice Integration}
In this work, we argue that expert advice is most beneficial when the agent is uncertain about the state or the risk of its actions. Leveraging the uncertainty measures introduced in Section~\ref{sec:uncertainty}, we design a query mechanism that triggers advice under uncertainty, employs a commitment–cooldown strategy to prevent over-reliance, and integrates expert guidance into training to accelerate policy improvement.

\subsubsection{Adaptive Query Mechanism}
A central challenge in leveraging uncertainty for advice queries is determining when uncertainty is high enough. Fixed thresholds are impractical because uncertainty magnitudes are not known in advance and evolve with the agent’s policy: thresholds set too high suppress valuable guidance during early training, while thresholds set too low risk over-reliance on the expert. To address this, we introduce an adaptive thresholding mechanism that evolves with training dynamics. Unlike heuristic or static thresholds used in prior work~\cite{hester2018deep, da2020uncertainty}, our approach adapts to recent uncertainty statistics, enabling consistent triggering of advice across different training phases.  

Epistemic and aleatoric uncertainties are stored in separate buffers 
that maintain a moving window of past values. At each step $t$, we derive an adaptive threshold by comparing the current uncertainty to the empirical $\beta$-percentile of its buffer distribution. Formally, the thresholds $T^{\text{epi}}_t$ and $T^{\text{ale}}_t$ are defined in 
Eq.~\ref{eq:adaptive_thresholds}, where $\mathrm{prc}_\beta(\cdot)$ denotes the $\beta$-percentile operator applied to the corresponding buffer of past values.
\begin{equation}
    T^{\text{epi}}_t = \mathrm{prc}_\beta\!\left(\{ U^{\text{epi}}_{t'} \}_{t'<t} \right), 
    \quad T^{\text{ale}}_t = \mathrm{prc}_\beta\!\left(\{ U^{\text{ale}}_{t'} \}_{t'<t} \right)
    \label{eq:adaptive_thresholds}
\end{equation}
If either epistemic or aleatoric uncertainty exceeds its respective threshold, the state is classified as high-uncertainty and triggers an advice request as defined in Eq.~\eqref{eq:advice_decision_1}, where $\mathds{1}[\cdot]$ denotes the indicator function.
\begin{equation}
\phi_t = \mathds{1}\!\left[ U^{\text{epi}}(o_t) > T^{\text{epi}}_t \;\lor\; U^{\text{ale}}(o_t,a^*) > T^{\text{ale}}_t \right]
\label{eq:advice_decision_1}
\end{equation}
This dynamic mechanism reflects the demands of \gls{ad}: even if the agent has experienced similar situations, high aleatoric uncertainty (e.g., an occluded intersection) still warrants caution, while high epistemic uncertainty in a less familiar road layout justifies consulting the expert.  

\subsubsection{Commitment–Cooldown Strategy}
Most prior work~\cite{da2020uncertainty, hoel2023ensemble} neglects the temporal regulation of expert advice, typically restricting advice to a single step. This has two drawbacks: (i) sparse corrections provide little opportunity to internalize the expert’s policy (e.g., a one-step lane-change correction does not expose the full maneuver), and (ii) unrestricted queries risk over-reliance and rapid exhaustion of the advice budget.

We address these issues with a commitment–cooldown strategy. During commitment, once triggered, the agent follows the expert for several consecutive steps, providing richer temporal context and exposure to coherent expert behavior. The subsequent cooldown blocks new advice requests, enforcing independent operation.

In addition to fixed commitment and cooldown periods, we propose a stochastic risk-sensitive heuristic for early termination of the commitment phase, inspired by the policy switching mechanism in~\cite{kurenkov2019ac}. The agent disengages from the expert once there is sufficient statistical evidence ($P_{\text{imp}}$) that its own policy achieves higher safety-adjusted returns. We quantify this evidence as the probability that the agent’s optimal action $a^*$ outperforms the expert’s action $a_E$, as defined in Eq.~\eqref{eq:p_imp}, where $X_a$ denotes a random variable approximating the safety-adjusted return of action $a$.
\begin{equation}
P_{\text{imp}} = P(X_{a^*} > X_{a_E})
\label{eq:p_imp}
\end{equation}
For each action $a$, every ensemble head $h$ produces a \gls{cvar} estimate $\operatorname{CVaR}^{(h)}_\alpha(o_t,a)$. We treat the resulting set $\{\operatorname{CVaR}^{(h)}_\alpha(o_t,a)\}_{h=1}^{N_H}$ as samples from the random variable $X_a$ and approximate it by a Gaussian distribution with empirical mean $\mu_{X_a}$ and variance $\sigma^2_{X_a}$.
To compare the optimal action $a^*$ with the expert’s action $a_E$, we define the difference $\Delta = X_{a^*} - X_{a_E}.$
Under the Gaussian approximation of $X_{a^*}$ and $X_{a_E}$, the difference $\Delta$ is also Gaussian, as given in Eq.~\eqref{eq:gauss_diff}.
\begin{equation}
\Delta \sim \mathcal{N}\,\big(\mu_{X_{a^*}} - \mu_{X_{a_E}}, \; \sigma_{X_{a^*}}^2 + \sigma_{X_{a_E}}^2 \big)
\label{eq:gauss_diff}
\end{equation}
Accordingly, $P_{\text{imp}} = P(\Delta > 0)$ is evaluated using the Gaussian cumulative distribution function. Early stopping of the commitment phase is triggered whenever $P_{\text{imp}} > \lambda \cdot \rho^{t_c}$, where $\lambda$ is a base confidence, $\rho \in (0,1)$ is a decay factor, and $t_c$ is the number of elapsed commitment steps. Since $\lambda \cdot \rho^{t_c}$ decreases monotonically with $t_c$, the confidence threshold lowers over time, making disengagement increasingly likely the longer advice is followed. 

\subsubsection{Integration in Policy Learning}
We integrate expert advice directly into the learning process by maintaining a shared replay buffer that stores both agent experiences and expert advice, leveraging the off-policy nature of \gls{iqn}. This contrasts with approaches that separate expert and agent buffers~\cite{peng2023learning}, which may keep expert samples disproportionately long and risk anchoring the policy to the expert’s behavior. In our design, the frequency of expert-generated transitions in the replay buffer decreases naturally as the agent becomes more confident and requests advice less frequently in similar states, better reflecting the agent’s evolving competence. 

Furthermore, since advice queries are triggered by uncertainty, the ratio of expert to agent samples adjusts automatically: when the agent is frequently uncertain, expert actions fill the buffer; when the agent is confident, its own experience dominates. This dynamic balance emerges naturally without imposing artificial ratios. Finally, the commitment-cooldown strategy further enriches the replay buffer with coherent trajectories rather than isolated corrections. By committing to short horizons of consecutive expert actions, the agent is exposed to complete maneuvers that provide more meaningful training signals. The subsequent cooldown enforces independence and prevents over-reliance on the expert.

We also avoid reward forcing~\cite{chekroun2023gri,peng2023learning}, where expert actions are artificially assigned high rewards. Such strategies implicitly assume that the expert is optimal, which is unrealistic in practice. Our expert may be suboptimal, and we want the learner to retain the ability to surpass it. By evaluating expert actions under the same reward function as the agent, the framework benefits from guidance without biasing the policy toward potentially flawed behavior.

Together, these choices ensure that expert knowledge is integrated in a balanced and principled way: leveraged when uncertainty is high, gradually fading as the agent becomes more competent, and always evaluated under the same reward function. This makes advice an accelerator for learning rather than a lasting dependency, embedding safety within the policy without limiting long-term autonomy.
\section{experimental Setup}
\label{sec:experiments}
This section details the experimental setup, including the \gls{rl} agent design, traffic scenarios, and the utilized expert. We also outline baselines, ablations, and evaluation metrics to enable a fair comparison of performance.

\subsection{RL Agent Description}
The \gls{rl} agent operates on a multimodal observation space that integrates complementary sensors and state information. The primary perception inputs consist of a frontal RGB camera image with a resolution of $128 \times 128$ pixels and a LiDAR point cloud projected into a $128 \times 128$ bird's-eye-view grid. To support goal-directed navigation, the LiDAR grid map is augmented with a reference route toward the target destination. The observation space also includes vehicle states, namely longitudinal and lateral velocities and their respective accelerations.

To encode the observation, we employ modality-specific encoders: a CNN for RGB images, a CNN for LiDAR grid maps, and an MLP for vehicle kinematics. Their outputs are fused into a shared latent representation that captures spatial and state information. Based on this representation, the policy head selects a cruise-control acceleration command from a finite action set uniformly spaced between maximum braking and maximum throttle. Expert actions are projected onto the same action space by selecting the closest command, while steering is handled by a route-following controller.

\subsection{Traffic Scenarios} 
In this work, we focus on urban driving tasks where an autonomous agent must safely approach and traverse unsignalized intersections. These intersections are among the most safety-critical components of road networks, as they lack explicit right-of-way rules and require implicit negotiation with surrounding traffic~\cite{al2024autonomous}. Although our framework is applicable to a wide range of driving scenarios, this work evaluates it in unsignalized intersections, a particularly challenging setting for evaluating our expert-guided \gls{rl} agents.

Traffic scenarios are generated using CARLA~\cite{dosovitskiy2017carla} across randomized configurations of traffic vehicles placed in multiple T-junctions and four-way intersections. To encourage generalization, we randomize traffic attributes such as vehicle geometry, speed, and lateral offset. For evaluation, we construct a hold-out set of one unseen T-junction and two unseen four-way intersections, ensuring that performance is assessed on layouts not encountered during training.

As an expert policy, we employ CARLA’s \gls{tm}, a rule-based system for controlling vehicle behavior. While the \gls{tm} has privileged access to the environment state, including road topology, and the state of all traffic vehicles, the framework itself is not restricted to this expert and only requires an action proposal from an external policy. Furthermore, the expert is required only during training and is neither required nor accessible during evaluation. 

\subsection{Baselines and Evaluation Metrics} 
We benchmark our proposed \gls{rl} algorithm against \gls{iqn}~\cite{dabney2018implicit}, a well-established and widely used distributional \gls{rl} baseline. In addition, we conduct a series of ablation studies to isolate the contribution of individual components of our approach, specifically evaluating the impact of the commitment–cooldown mechanism, the expert advice budget, and the stochastic early stopping strategy for the commitment mechanism. These analyses provide a clearer understanding of how each design choice influences the overall learning performance.

To ensure fair comparison, all agents are trained for an identical number of steps using the same network architectures and hyperparameter configurations. Each training run is repeated with three independent random seeds, and every trained policy is evaluated over three runs to account for the stochasticity inherent in CARLA. This setup follows the evaluation methodology outlined in~\cite{jaeger2025carl}.

Evaluation is performed on a hold-out set of intersection scenarios with varying traffic densities. Traffic density is the ratio of the number of active traffic vehicles to the environment's maximum capacity. Results are reported as mean and standard deviation over all training seeds and evaluation runs. Performance is measured using cumulative episode reward ($ER$) as well as driving-specific metrics: success rate ($SR$), failure rate ($FR$), the sum of off-road, collision, and timeout rates, and route progress ($RP$). To ensure robust evaluation, we use RLiable~\cite{agarwal2021deep} to report aggregate metrics such as interquartile mean (IQM) and optimality gap, which better capture performance variability.
\begin{figure}[t]
    \centering
    \includegraphics[width=0.75\linewidth,trim=8 10 8 0,clip]{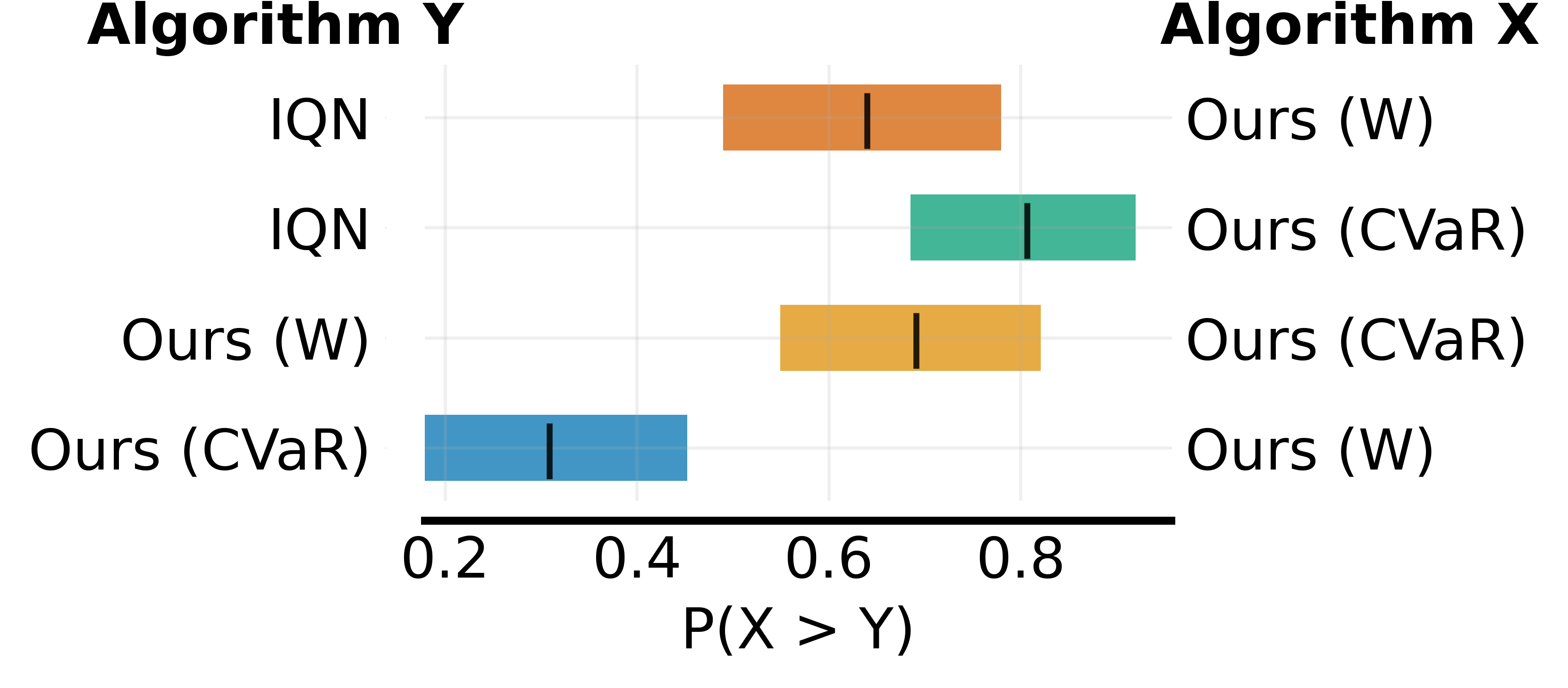}
    \caption{Probability of improvement~\cite{agarwal2021deep}, quantifying the likelihood that algorithm X outperforms algorithm Y.}
    \label{fig:prob}
\end{figure}
\section{Evaluation}
Table~\ref{tab:main_results} reports ablation studies evaluating our methodology across traffic densities. Introducing \emph{commitment and cooldown periods} proves essential, as they ensure more consistent expert guidance. The baselines~\cite{wei2025agent, da2020uncertainty} without this mechanism correspond to $(1,1)$, where success reaches $0.67$ at density $0.75$ and $0.53$ at density $1.0$. A suitable period of $(5,5)$ raises success to $0.74$ and $0.61$, respectively, while reducing failures from $0.33$ to $0.26$ at density $0.75$ and from $0.47$ to $0.39$ at density $1.0$. This demonstrates that temporal consistency during training stabilizes policy learning.  

The expert budget plays a central role, as it represents the fraction of training during which the agent can receive advice. A limited budget of 25\% already enhances performance, while 50\% yields the best balance between advice and independent learning. In contrast, allocating 75\% induces excessive reliance on the expert and degrades generalization. 
At the balanced $50\%$, success improves from $0.67$ for \gls{iqn} to $0.74$ at density $0.75$ and from $0.55$ to $0.61$ at density $1.0$.

When comparing epistemic uncertainty formulations, both Wasserstein and \gls{cvar} follow similar trends across commitment settings and budgets. However, \gls{cvar} consistently achieves higher success (e.g., $0.61$ vs. $0.58$ at density $1.0$) by emphasizing risk-sensitive quantiles and filtering distributional noise. The stochastic early-stop mechanism shows a nuanced effect. At the optimal $50\%$ budget, early stopping causes minor degradation, whereas at larger budgets it mitigates overreliance. At density $0.75$ with $75\%$ budget, success increases from $0.64$ to $0.72$, indicating that stochastic early-stop can reduce overreliance during longer training schedules and expert budgets. 

Overall, our best configuration, CVaR-based uncertainty with $(5,5)$ commitment-cooldown and a $50\%$ expert budget, delivers consistent improvements across all densities. Compared to IQN, it yields $5$--$7\%$ gains in success rate, while lowering failures in challenging dense traffic. These results confirm the effectiveness of combining structured expert advice with risk-sensitive uncertainty estimation for the evaluated intersection-navigation setting. 

\begin{figure}[t]
    \centering
    \includegraphics[width=1.0\linewidth,trim=8 8 8 2,clip]{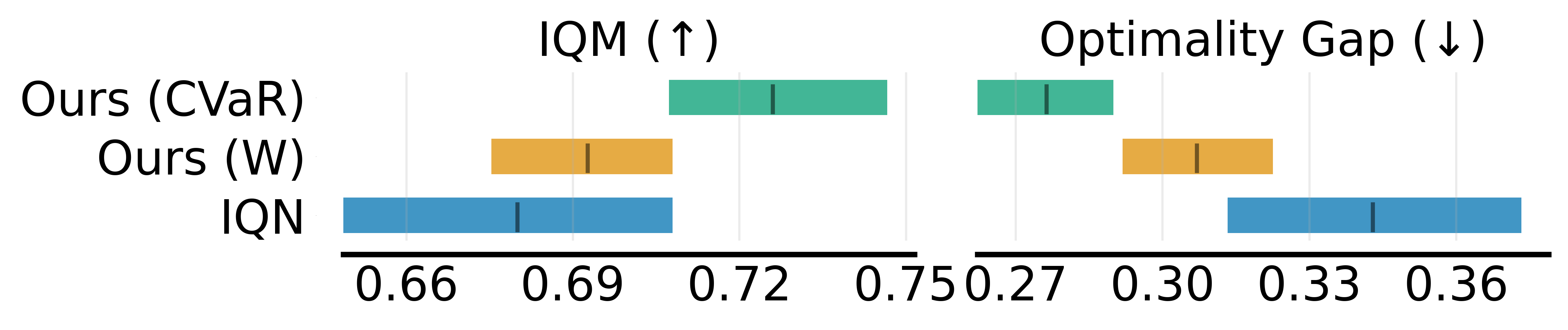}
    \caption{Interquartile mean (IQM) and optimality gap~\cite{agarwal2021deep}, quantifying the statistical stability of a policy.}
    \label{fig:iqm}
\end{figure}

Furthermore, RLiable analysis confirms these findings. Fig.~\ref{fig:iqm} shows that our best variants raise IQM from $0.66$ (\gls{iqn}) to $0.72$ (\gls{cvar}) and reduce the optimality gap from $0.33$ to $0.27$. Pairwise improvement probabilities (Fig.~\ref{fig:prob}) further highlight consistency: CVaR outperforms IQN with probability $>0.75$, while Wasserstein remains above $0.6$. Thus, our best configurations deliver not only higher mean performance but also more reliable learning outcomes.

Having established the effectiveness of our best configurations through both direct metrics and RLiable analysis, we finally evaluate aleatoric uncertainty as a runtime safety guard. At inference time, the agent triggers a deceleration-to-stop maneuver whenever the current uncertainty exceeds the $90^{\text{th}}$ percentile threshold of uncertainty values faced on a validation set. Under density $0.75$, this mechanism achieves the highest success rates of $0.76$ with CVaR-based epistemic uncertainty and $0.71$ with the Wasserstein variant, surpassing all previous configurations. These findings suggest that runtime uncertainty monitoring complements expert advice regulation, providing an additional layer of robustness when the agent encounters rare failure cases.
\begin{table*}[t]
\centering
\caption{Ablation results in CARLA traffic scenarios across traffic densities. Results compare IQN with our method under different commitment–cooldown periods, expert budgets, and uncertainty formulations.}
\label{tab:main_results}
\renewcommand{\arraystretch}{1.3}
\resizebox{\linewidth}{!}{
\setlength{\tabcolsep}{4pt}
\begin{tabular}{p{0.9cm}cccc|cccc}
\toprule
 & \multicolumn{4}{c}{Traffic Density 0.75} & \multicolumn{4}{c}{Traffic Density 1.0} \\
\cmidrule(lr){2-5} \cmidrule(lr){6-9}
\textbf{Setup} 
& SR $\uparrow$ & FR $\downarrow$ & RP $\uparrow$ & ER $\uparrow$ 
& SR $\uparrow$ & FR $\downarrow$ & RP $\uparrow$ & ER $\uparrow$ \\
\midrule
IQN   & $0.670 \pm 0.054$ & $0.330 \pm 0.058$ & $0.666\pm0.025$ &$0.549 \pm 0.025$ 
      & $0.553 \pm 0.065$& $0.440\pm 0.133$ & $0.608 \pm 0.065$ & $0.327 \pm 0.085$\\
\midrule
\multicolumn{9}{c}{\textbf{Ablation of commitment and cooldown periods with $U^{\text{epi}}_{\text{CVaR}}$ and expert budget = 50\% of training steps}} \\
\midrule
(1, 1)   & $0.673\pm0.043$ &$0.327 \pm 0.048$ & $0.662\pm0.026$& $0.536\pm0.033$ 
         & $0.527\pm0.056$ & $0.472 \pm 0.063$&$0.558\pm0.026$ &$0.297\pm0.038$ \\

(5, 5) & $0.735\pm0.048$ & $0.264\pm0.052$ & $0.695\pm0.022$ & $0.590\pm0.032$ 
       & $0.607\pm0.027$& $0.392\pm0.036$ & $0.607\pm0.026$ & $0.362\pm0.051$\\

(10, 10) &$0.696\pm0.088$ & $0.303 \pm 0.088$& $0.681\pm 0.034$ & $0.572\pm0.056$ 
         & $0.593\pm0.097$ & $0.407 \pm 0.111$ & $0.602\pm0.050$& $0.326\pm0.067$ \\

(20, 10) &$0.712\pm0.054$ & $0.288 \pm 0.054$& $0.694\pm0.018$& $0.581\pm0.030$ 
         & $0.600\pm0.033$& $0.400 \pm 0.048$ & $0.617\pm0.019$&$0.358\pm0.030$ \\

(10, 20) & $0.659 \pm 0.073$ & $0.341 \pm 0.080$ & $0.644 \pm 0.034$ & $0.520 \pm 0.045$ 
         & $0.534 \pm 0.075$ & $0.466 \pm 0.081$ & $0.554 \pm 0.043$ & $0.275 \pm 0.056$ \\
\midrule
\multicolumn{9}{c}{\textbf{Ablation of expert budget with $U^{\text{epi}}_{\text{CVaR}}$ and best commitment and cooldown periods = (5, 5)}} \\
\midrule
25\%& $0.694 \pm 0.053$ & $0.306 \pm 0.058$ & $0.671 \pm 0.032$ & $0.511 \pm 0.027$ 
    & $0.577 \pm 0.086$ & $0.423 \pm 0.097$ & $0.578 \pm 0.049$ & $0.308 \pm 0.042$ \\
75\%& $0.642 \pm 0.038$ & $0.358 \pm 0.043$ & $0.644 \pm 0.025$ & $0.510 \pm 0.035$ 
    & $0.471 \pm 0.038$ & $0.529 \pm 0.053$ & $0.522 \pm 0.029$ & $0.236 \pm 0.020$ \\
\midrule

\multicolumn{9}{c}{\textbf{Ablation of commitment and cooldown periods with $U^{\text{epi}}_{\text{W}}$ and expert budget = 50\% of training steps}} \\
\midrule
(1, 1) & $0.643 \pm 0.074$ & $0.357 \pm 0.085$ & $0.646 \pm 0.038$ & $0.498 \pm 0.055$
       & $0.476 \pm 0.131$ & $0.524 \pm 0.141$ & $0.546 \pm 0.083$ & $0.228 \pm 0.143$ \\
(5, 5) & $0.726 \pm 0.040$ & $0.274 \pm 0.048$ & $0.687 \pm 0.027$ & $0.569 \pm 0.024$
       & $0.578 \pm 0.049$ & $0.422 \pm 0.053$ & $0.575 \pm 0.026$ & $0.317 \pm 0.033$ \\
(10, 10)& $0.679 \pm 0.036$ & $0.321 \pm 0.045$ & $0.668 \pm 0.022$ & $0.516 \pm 0.031$
        & $0.606 \pm 0.052$ & $0.394 \pm 0.069$ & $0.606 \pm 0.033$ & $0.339 \pm 0.029$ \\
\midrule
\multicolumn{9}{c}{\textbf{Ablation of expert budget with $U^{\text{epi}}_{\text{W}}$ and best commitment and cooldown periods = (10, 10)}} \\
\midrule
25\% & $0.647 \pm 0.080$ & $0.353 \pm 0.086$ & $0.654 \pm 0.037$ & $0.511 \pm 0.050$
     & $0.509 \pm 0.103$ & $0.491 \pm 0.114$ & $0.563 \pm 0.049$ & $0.260 \pm 0.085$ \\
75\% & $0.622 \pm 0.074$ & $0.378 \pm 0.077$ & $0.643 \pm 0.037$ & $0.520 \pm 0.064$ 
     & $0.440 \pm 0.085$ & $0.560 \pm 0.097$ & $0.531 \pm 0.046$ & $0.222 \pm 0.089$ \\
\midrule
\multicolumn{9}{c}{\textbf{Impact of stochastic commitment early stop with $U^{\text{epi}}_{\text{CVaR}}$ and best commitment and cooldown periods = (5, 5)}} \\
\midrule
50\% & $0.714 \pm 0.060$ & $0.286 \pm 0.065$ & $0.691 \pm 0.034$ & $0.554 \pm 0.024$
     & $0.604 \pm 0.078$ & $0.396 \pm 0.093$ & $0.611 \pm 0.052$ & $0.333 \pm 0.054$ \\
75\% & $0.721 \pm 0.041$ & $0.279 \pm 0.045$ & $0.688 \pm 0.024$ & $0.580 \pm 0.048$
     & $0.593 \pm 0.069$ & $0.407 \pm 0.082$ & $0.599 \pm 0.046$ & $0.335 \pm 0.086$ \\
\midrule
\multicolumn{9}{c}{\textbf{Impact of stochastic commitment early stop with $U^{\text{epi}}_{\text{W}}$ and best commitment and cooldown periods = (10, 10)}} \\
\midrule
50\% & $0.671 \pm 0.052$ & $0.329 \pm 0.064$ & $0.668 \pm 0.034$ & $0.542 \pm 0.044$
     & $0.528 \pm 0.085$ & $0.472 \pm 0.098$ & $0.575 \pm 0.048$ & $0.295 \pm 0.068$ \\
75\% & $0.647 \pm 0.041$ & $0.353 \pm 0.056$ & $0.652 \pm 0.028$ & $0.530 \pm 0.045$
     & $0.481 \pm 0.074$ & $0.519 \pm 0.080$ & $0.533 \pm 0.042$ & $0.235 \pm 0.058$ \\
\midrule
\multicolumn{9}{c}{\textbf{Impact of using aleatoric uncertainty as an inference safety guard}} \\
\midrule
$U^{\text{epi}}_{\text{CVaR}}$  & $0.761\pm0.049$& $0.239 \pm 0.043$ &$0.714\pm0.019$ & $0.571\pm0.025$
& $0.672\pm0.047$& $0.328 \pm 0.053$&$0.645\pm0.030$ &$0.399\pm0.033$ \\
$U^{\text{epi}}_{\text{W}}$ & $0.714\pm0.030$ & $0.286 \pm 0.033$&  $0.688\pm0.014$&$0.499\pm0.021$ 
& $0.678\pm0.049$& $0.322 \pm 0.051$& $0.650\pm0.028$ & $0.378\pm0.031$\\
\bottomrule
\end{tabular}
}
\vspace{-0.1cm}
\end{table*}

\section{Conclusion}
This work addressed the gap in safe exploration by introducing an uncertainty-aware framework that triggers advice when epistemic or aleatoric uncertainty exceeds adaptive thresholds and regulates its use with a commitment–cooldown strategy, exposing the agent to coherent expert trajectories without inducing long-term dependence. Experiments in CARLA traffic scenarios demonstrated consistent gains over IQN, with the best configuration (CVaR-based uncertainty, and a $50\%$ advice budget with $(5,5)$ periods) achieving $5$–$7\%$ higher success and lower failures across all traffic densities. In addition, we showed that aleatoric uncertainty can serve as an inference-time safety guard, further improving success under dense traffic and providing robustness against rare failure cases. Future work will explore noisy, partially observable, and multi-expert settings, where specialized experts can be queried selectively, enhancing flexibility and broadening applicability to diverse and safety-critical driving contexts.

\section*{ACKNOWLEDGMENT}
The research leading to these results is funded by the German Federal Ministry for Economic Affairs and Energy within the project “Safe AI Engineering – Sicherheitsargumentation befähigendes AI Engineering über den gesamten Lebenszyklus einer KI-Funktion". The authors would like to thank the consortium for the successful cooperation.

{
    \bibliographystyle{IEEEtran}
    \bibliography{references}

@IEEEtranBSTCTL{IEEEexample:BSTcontrol,
CTLuse_forced_etal       = "yes",
CTLmax_names_forced_etal = "6",
CTLnames_show_etal       = "4" }

@article{chen2024end,
  title={End-to-end autonomous driving: Challenges and frontiers},
  author={Chen, Li and Wu, Penghao and Chitta, Kashyap and Jaeger, Bernhard and Geiger, Andreas and Li, Hongyang},
  journal={IEEE Transactions on Pattern Analysis and Machine Intelligence},
  year={2024},
}

@article{gu2024review,
  title={A Review of Safe Reinforcement Learning: Methods, Theories and Applications},
  author={Gu, Shangding and Yang, Long and Du, Yali and Chen, Guang and Walter, Florian and Wang, Jun and Knoll, Alois},
  journal={IEEE Transactions on Pattern Analysis and Machine Intelligence},
  year={2024},
}

@article{wachi2023safe,
  title={Safe exploration in reinforcement learning: A generalized formulation and algorithms},
  author={Wachi, Akifumi and Hashimoto, Wataru and Shen, Xun and Hashimoto, Kazumune},
  journal={Advances in Neural Information Processing Systems},
  year={2023}
}

@book{sutton1998reinforcement,
  title={Reinforcement learning: An introduction},
  author={Sutton, Richard S and Barto, Andrew G and others},
  year={1998},
  publisher={MIT press Cambridge}
}

@article{dulac2021challenges,
  title={Challenges of real-world reinforcement learning: definitions, benchmarks and analysis},
  author={Dulac-Arnold, Gabriel and Levine, Nir and Mankowitz, Daniel J and Li, Jerry and Paduraru, Cosmin and Gowal, Sven and Hester, Todd},
  journal={Machine Learning},
  year={2021},
}

@article{al2024autonomous,
  author={Al-Sharman, Mohammad and Edes, Luc and Sun, Bert and Jayakumar, Vishal and Tahir, Hasan and Daoud, Mohamed A. and Emran, Bara J. and Rayside, Derek and Melek, William},
  journal={IEEE Transactions on Automation Science and Engineering}, 
  title={Autonomous Driving at Unsignalized Intersections: A Review of Decision-Making Challenges and Reinforcement Learning-Based Solutions}, 
  year={2026},
  doi={10.1109/TASE.2025.3646982}}

@inproceedings{dosovitskiy2017carla,
  title={CARLA: An open urban driving simulator},
  author={Dosovitskiy, Alexey and Ros, German and Codevilla, Felipe and Lopez, Antonio and Koltun, Vladlen},
  booktitle={Conference on robot learning},
  year={2017},
}

@inproceedings{dabney2018implicit,
  title={Implicit quantile networks for distributional reinforcement learning},
  author={Dabney, Will and Ostrovski, Georg and Silver, David and Munos, R{\'e}mi},
  booktitle={International conference on machine learning},
  year={2018},
}

@InProceedings{jaeger2025carl, 
	author = {Bernhard Jaeger and Daniel Dauner and Jens Beißwenger and Simon Gerstenecker and Kashyap Chitta and Andreas Geiger}, 
	title = {CaRL: Learning Scalable Planning Policies with Simple Rewards}, 
	booktitle = {Proc. of the Conf. on Robot Learning (CoRL)}, 
	year = {2025}, 
}

@article{mo2021safe,
  title={Safe reinforcement learning for autonomous vehicle using monte carlo tree search},
  author={Mo, Shuojie and Pei, Xiaofei and Wu, Chaoxian},
  journal={IEEE Transactions on Intelligent Transportation Systems},
  year={2021},
}

@article{hu2024long,
  title={Long-and Short-Term Constraint-Driven Safe Reinforcement Learning for Autonomous Driving},
  author={Hu, Xuemin and Chen, Pan and Wen, Yijun and Tang, Bo and Chen, Long},
  journal={IEEE Transactions on Systems, Man, and Cybernetics: Systems},
  year={2026},
  publisher={IEEE}
}

@article{chekroun2023gri,
  title={Gri: General reinforced imitation and its application to vision-based autonomous driving},
  author={Chekroun, Raphael and Toromanoff, Marin and Hornauer, Sascha and Moutarde, Fabien},
  journal={Robotics},
  year={2023},
}

@inproceedings{hester2018deep,
  title={Deep q-learning from demonstrations},
  author={Hester, Todd and Vecerik, Matej and Pietquin, Olivier and Lanctot, Marc and Schaul, Tom and Piot, Bilal and Horgan, Dan and Quan, John and Sendonaris, Andrew and Osband, Ian and others},
  booktitle={Proceedings of the AAAI conference on artificial intelligence},
  year={2018}
}

@article{hoel2023ensemble,
  title={Ensemble quantile networks: Uncertainty-aware reinforcement learning with applications in autonomous driving},
  author={Hoel, Carl-Johan and Wolff, Krister and Laine, Leo},
  journal={IEEE Transactions on Intelligent Transportation Systems},
  year={2023},
}

@article{osband2016deep,
  title={Deep exploration via bootstrapped DQN},
  author={Osband, Ian and Blundell, Charles and Pritzel, Alexander and Van Roy, Benjamin},
  journal={Advances in neural information processing systems},
  volume={29},
  year={2016}
}

@inproceedings{lockwood2022review,
  title={A review of uncertainty for deep reinforcement learning},
  author={Lockwood, Owen and Si, Mei},
  booktitle={Proceedings of the AAAI Conference on Artificial Intelligence and Interactive Digital Entertainment},
  year={2022}
}

@article{lakshminarayanan2017simple,
  title={Simple and scalable predictive uncertainty estimation using deep ensembles},
  author={Lakshminarayanan, Balaji and Pritzel, Alexander and Blundell, Charles},
  journal={Advances in neural information processing systems},
  volume={30},
  year={2017}
}

@inproceedings{da2020uncertainty,
  title={Uncertainty-aware action advising for deep reinforcement learning agents},
  author={Da Silva, Felipe Leno and Hernandez-Leal, Pablo and Kartal, Bilal and Taylor, Matthew E},
  booktitle={Proceedings of the AAAI conference on artificial intelligence},
  year={2020}
}

@article{kurenkov2019ac,
  title={Ac-teach: A bayesian actor-critic method for policy learning with an ensemble of suboptimal teachers},
  author={Kurenkov, Andrey and Mandlekar, Ajay and Martin-Martin, Roberto and Savarese, Silvio and Garg, Animesh},
  journal={arXiv preprint arXiv:1909.04121},
  year={2019}
}

@inproceedings{kamran2021minimizing,
  title={Minimizing safety interference for safe and comfortable automated driving with distributional reinforcement learning},
  author={Kamran, Danial and Engelgeh, Tizian and Busch, Marvin and Fischer, Johannes and Stiller, Christoph},
  booktitle={2021 IEEE/RSJ international conference on intelligent robots and systems (IROS)},
  year={2021},
}

@article{liu2024automatic,
  title={An automatic driving trajectory planning approach in complex traffic scenarios based on integrated driver style inference and deep reinforcement learning},
  author={Liu, Yuchen and Diao, Shuzhen},
  journal={PLoS one},
  year={2024},
  publisher={Public Library of Science San Francisco, CA USA}
}

@article{wang2023autonomous,
  title={Autonomous driving based on approximate safe action},
  author={Wang, Xuesong and Zhang, Jiazhi and Hou, Diyuan and Cheng, Yuhu},
  journal={IEEE Transactions on Intelligent Transportation Systems},
  year={2023},
}

@INPROCEEDINGS{du2023reinforcement,
  author={Du, Desong and Han, Shaohang and Qi, Naiming and Ammar, Haitham Bou and Wang, Jun and Pan, Wei},
  booktitle={2023 IEEE International Conference on Robotics and Automation (ICRA)}, 
  title={Reinforcement Learning for Safe Robot Control using Control Lyapunov Barrier Functions}, 
  year={2023},
  doi={10.1109/ICRA48891.2023.10160991}
}

@article{tan2023value,
  title={Value functions are control barrier functions: Verification of safe policies using control theory},
  author={Tan, Daniel CH and Acero, Fernando and McCarthy, Robert and Kanoulas, Dimitrios and Li, Zhibin},
  journal={arXiv preprint arXiv:2306.04026},
  year={2023}
}

@inproceedings{tan2024safe,
  title={Safe Value Functions: Learned Critics as Hard Safety Constraints},
  author={Tan, Daniel CH and McCarthy, Robert and Acero, Fernando and Delfaki, Andromachi Maria and Li, Zhibin and Kanoulas, Dimitrios},
  booktitle={2024 IEEE 20th International Conference on Automation Science and Engineering (CASE)},
  year={2024},
}

@inproceedings{bernhard2019addressing,
  title={Addressing inherent uncertainty: Risk-sensitive behavior generation for automated driving using distributional reinforcement learning},
  author={Bernhard, Julian and Pollok, Stefan and Knoll, Alois},
  booktitle={2019 IEEE Intelligent Vehicles Symposium (IV)},
  year={2019},
}

@inproceedings{choi2025enhancing,
  title={Enhancing Autonomous Driving with Pre-trained Imitation and Reinforcement Learning},
  author={Choi, Jeong-Hwan and Kim, Dong-han and Yoo, Ji-Sang and Kim, Beom-Joon and Hwang, Jun-Tae},
  booktitle={2025 International Conference on Electronics, Information, and Communication (ICEIC)},
  year={2025},
}

@article{pfeiffer2018reinforced,
  title={Reinforced imitation: Sample efficient deep reinforcement learning for mapless navigation by leveraging prior demonstrations},
  author={Pfeiffer, Mark and Shukla, Samarth and Turchetta, Matteo and Cadena, Cesar and Krause, Andreas and Siegwart, Roland and Nieto, Juan},
  journal={IEEE Robotics and Automation Letters},
  year={2018},
}

@article{gao2025ril,
  title={IN-RIL: Interleaved Reinforcement and Imitation Learning for Policy Fine-Tuning},
  author={Gao, Dechen and Wang, Hang and Zhou, Hanchu and Ammar, Nejib and Mishra, Shatadal and Moradipari, Ahmadreza and Soltani, Iman and Zhang, Junshan},
  journal={arXiv preprint arXiv:2505.10442},
  year={2025}
}

@article{peng2023learning,
  title={Learning from active human involvement through proxy value propagation},
  author={Peng, Zhenghao Mark and Mo, Wenjie and Duan, Chenda and Li, Quanyi and Zhou, Bolei},
  journal={Advances in neural information processing systems},
  year={2023}
}

@article{huang2024human,
  title={Human as AI mentor: Enhanced human-in-the-loop reinforcement learning for safe and efficient autonomous driving},
  author={Huang, Zilin and Sheng, Zihao and Ma, Chengyuan and Chen, Sikai},
  journal={Communications in Transportation Research},
  year={2024},
}

@article{wang2023safe,
  title={Safe reinforcement learning for automated vehicles via online reachability analysis},
  author={Wang, Xiao and Althoff, Matthias},
  journal={IEEE Transactions on Intelligent Vehicles},
  year={2023},
}

@inproceedings{thumm2023reducing,
  title={Reducing safety interventions in provably safe reinforcement learning},
  author={Thumm, Jakob and Pelat, Guillaume and Althoff, Matthias},
  booktitle={2023 IEEE/RSJ International Conference on Intelligent Robots and Systems (IROS)},
  year={2023},
}

@article{wang2023learning,
  title={Learning to Recover for Safe Reinforcement Learning},
  author={Wang, Haoyu and Yuan, Xin and Ren, Qinqing},
  journal={arXiv preprint arXiv:2309.11907},
  year={2023}
}

@inproceedings{peng2022safe,
  title={Safe driving via expert guided policy optimization},
  author={Peng, Zhenghao and Li, Quanyi and Liu, Chunxiao and Zhou, Bolei},
  booktitle={Conference on Robot Learning},
  year={2022},
}

@article{xue2023guarded,
  title   = {Guarded Policy Optimization with Imperfect Online Demonstrations},
  author  = {Zhenghai Xue and Zhenghao Peng and Quanyi Li and Zhihan Liu and Bolei Zhou},
  journal = {International Conference on Learning Representations},
  year    = {2023},
  url     = {https://openreview.net/forum?id=O5rKg7IRQIO}
}

@inproceedings{kelly2019hg, 
title={Hg-dagger: Interactive imitation learning with human experts}, 
author={Kelly, Michael and Sidrane, Chelsea and Driggs-Campbell, Katherine and Kochenderfer, Mykel J}, 
booktitle={2019 International Conference on Robotics and Automation (ICRA)}, 
year={2019}, 
}

@article{ilhan2021student, 
title={Student-initiated action advising via advice novelty}, 
author={Ilhan, Ercument and Gow, Jeremy and Perez, Diego}, 
journal={IEEE Transactions on Games}, 
year={2021}, 
}

@inproceedings{wei2025agent, 
title={Agent-Aware Training for Agent-Agnostic Action Advising in Deep Reinforcement Learning}, 
author={Wei, Yaoquan and Liu, Shunyu and Song, Jie and Zheng, Tongya and Chen, Kaixuan and Song, Mingli}, booktitle={Proceedings of the AAAI Conference on Artificial Intelligence}, 
year={2025} 
}

@article{bethell2024safe, 
title={Safe reinforcement learning in black-box environments via adaptive shielding}, 
author={Bethell, Daniel and Gerasimou, Simos and Calinescu, Radu and Imrie, Calum}, 
journal={arXiv preprint arXiv:2405.18180}, 
year={2024} 
}

@article{agarwal2021deep,
  title={Deep Reinforcement Learning at the Edge of the Statistical Precipice},
  author={Agarwal, Rishabh and Schwarzer, Max and Castro, Pablo Samuel and Courville, Aaron and Bellemare, Marc G},
  journal={NeurIPS},
  year={2021}
}
}

\end{document}